\documentclass[10pt,letterpaper]{article}

\usepackage{graphicx}
\usepackage{booktabs}
\usepackage{array}
\usepackage{xcolor}
\usepackage{cogsci}
\usepackage{amsmath}

\cogscifinalcopy 

\usepackage{pslatex}
\usepackage{apacite}
\usepackage{float} 

\begin{document}

\title{DPMT: Dual Process Multi-scale Theory of Mind Framework for Real-time Human-AI Collaboration}

\author{{\large \bf Xiyun Li (lixiyun2020@ia.ac.cn)}
\textsuperscript{1,2},\large \bf Yining Ding\textsuperscript{3},\large \bf Yuhua Jiang\textsuperscript{4},\large \bf Yunlong Zhao\textsuperscript{1},
\large \bf Runpeng Xie\textsuperscript{1}, \\
\large \bf Shuang Xu\textsuperscript{1},\large \bf Yuanhua Ni\textsuperscript{3},
\large \bf Yiqin Yang (yiqin.yang@ia.ac.cn)\textsuperscript{1*},\large 
\bf Bo Xu (xubo@ia.ac.cn)\textsuperscript{1,2*}  \\
  \textsuperscript{1}The Key Laboratory of Cognition and Decision Intelligence for Complex Systems, \\ Institute of Automation, Chinese Academy of Sciences, Beijing, 100190, China\\
  \textsuperscript{2}School of Future Technology, University of Chinese Academy of Sciences, Beijing, 101408, China \\
  \textsuperscript{3}Nankai University, Tianjin, 300350, China \\
  \textsuperscript{4}Tsinghua University, Beijing, 100084, China \\
  \small{*Corresponding author}}

\maketitle

\begin{abstract}
Real-time human-artificial intelligence (AI) collaboration is crucial yet challenging, especially when AI agents must adapt to diverse and unseen human behaviors in dynamic scenarios. Existing large language model (LLM) agents often fail to accurately model the complex human mental characteristics such as domain intentions, especially in the absence of direct communication. To address this limitation, we propose a novel dual process multi-scale theory of mind (DPMT) framework, drawing inspiration from cognitive science's dual process theory. Our DPMT framework incorporates a multi-scale theory of mind (ToM) module to facilitate robust human partner modeling through mental characteristic reasoning. Experimental results demonstrate that DPMT significantly enhances human-AI collaboration, and ablation studies further validate the contributions of our multi-scale ToM in the slow system.

\textbf{Keywords:} 
theory of mind; human-AI collaboration; dual process theory; intelligent agents; large language model
\end{abstract}

\section{Introduction}
In recent years, large language models (LLMs) have achieved expert-level performance across various domains, including conversational question-answering assistants~\cite{achiam2023gpt} and code generation~\cite{chen2021evaluating, li2023starcoder}. Building on the exceptional perception, comprehension, and reasoning capabilities of LLMs, LLM agents have rapidly emerged, attracting widespread attention. By integrating mechanisms such as memory and reflection, these agents demonstrate general-purpose abilities and robust generalization across tasks like operating system command execution~\cite{deng2024mind2web, he2024webvoyager}, mobile assistance~\cite{wang2024mobile, zhang2023appagent}, creative exploration in Minecraft~\cite{wang2023voyager}, and embodied intelligence scenarios~\cite{brohan2023can}.

However, achieving effective adaptation in complex collaborative tasks remains a significant challenge, which is critical for the broader application of LLM agents. Unlike tasks that an LLM agent can perform independently, collaborative tasks like \textit{Overcooked}~\cite{ghosttown2016overcooked} require the agent to work with diverse partners, including humans, to efficiently complete a series of complex sub-tasks—such as retrieving ingredients, cooking, and serving dishes within a limited time. The task list in \textit{Overcooked} is too intricate for a single agent to handle alone. Furthermore, human collaborators, constrained by bounded rationality, may lack complete domain knowledge, introducing additional instability and uncertainty in human-artificial intelligence (AI) collaboration. Thus, effectively modeling human cognition is key to improving adaptation and collaboration.

Recent research has focused on improving the adaptability and decision-making capabilities of LLM agents by facilitating communication between human players and AI agents in environments with diverse human partners \cite{liu2023llm,zhang2024mutual}. Leveraging their advanced natural language understanding, LLM agents can interpret human commands and formulate subsequent action plans~\cite{liu2023llm}. By adjusting their collaboration strategies through mutual communication with human partners, LLM agents significantly improve the overall collaboration performance of human-AI teams~\cite{zhang2024mutual}.

However, the improved communication alone does not fully address the challenges of real-time scenarios. Current LLM agents lack the human-like cognitive ability known as ``theory of mind" (ToM), which allows humans to understand and predict others' mental beliefs based on observed behaviors in social environment~\cite{astington1995theory}. This ability facilitates efficient collaboration in tasks without direct communication. The absence of ToM hinders LLM agents' performance in complex, real-time human-AI collaborative tasks. Although recent studies have explored ToM modeling to improve agent prediction~\cite{rabinowitz2018machine, li17mixture}, these approaches rely on high-quality trajectories and prior knowledge of partners. As a result, their interpretability and generalization are limited, restricting their applicability in real collaborative scenarios.

\begin{figure}[htb]
\begin{center}
\includegraphics[width=0.48\textwidth]{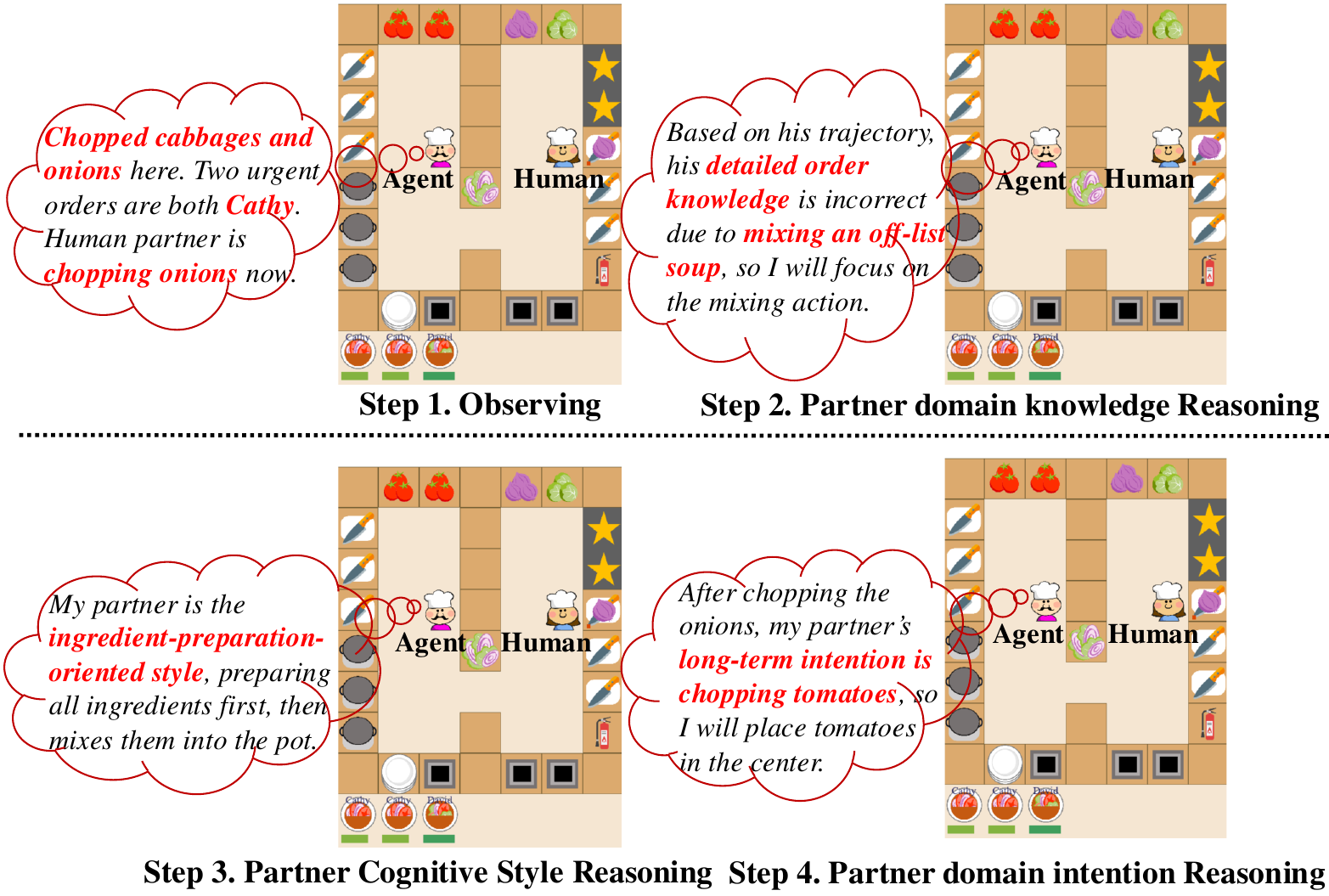}
\end{center}
\caption{A four-step example of cognitive ToM modeling for human partners in the collaborative task Overcooked. In Step 1, the agent observes the environment and the human partner’s recent behaviors. In Step 2, the agent understands the human trajectory and predicts the partner’s domain knowledge. In Step 3, the agent analyzes the partner’s behavioral preferences and categorizes his cognitive style based on the inferred domain knowledge. In Step 4, the agent infers the partner’s domain intention for effective disicion-making.}
\label{intro_tom}
\end{figure}
Inspired by dual process theory~\cite{vaisey2009motivation, lizardo2016dual}, we propose a \textit{cognitive dual process multi-scale theory of mind} (DPMT) to improve the interpretability and efficiency of human partner modeling in real-time human-AI collaboration. Our DPMT framework distinguishes between two decision-making systems for real-time human-AI collaboration: a \textit{fast system} for automatic decisions and a \textit{slow system} for modeling higher-level cognitive abilities. The core contribution of this work is the development of a \textit{multi-scale theory of mind module} to simulate the slow system for understanding human partners' behavioral trajectories and reasoning about their mental characteristics, facilitating more effective collaboration. This ToM process follows a three-tiered reasoning process, which progresses from domain knowledge to cognitive style, and ultimately to domain intention, as illustrated in Figure~\ref{intro_tom}. Experimental results from collaborative tasks in \textit{Overcooked} demonstrate the effectiveness of DPMT in improving real-time human-AI collaboration.

\subsection{Related Work}\paragraph{LLM agent.}Human-AI collaboration~\cite{puigwatch,gao2020joint} constitutes a highly challenging research direction, motivating extensive investigations into strategies for enhancing the effectiveness, efficiency, and adaptability of human-AI systems. Compared to traditional agents, LLM-based agents contain an inherent world model, making them highly effective in reasoning and facilitating interpretable interaction for complex tasks, such as human-AI collaboration~\cite{wu2024spring,liu2023llm}. Recent research \cite{zhang2024proagent} designs the LLM agent framework with a verifier on the \textit{Overcooked} environment, with knowledge and state as prompts to guide action planning. Recent work by \cite{zhang2023building} proposes a modular framework that enables embodied agents to communicate and collaborate effectively, facilitating the efficient accomplishment of multi-agent long-term collaboration tasks. 
\paragraph{Theory of Mind (ToM)} refers to the ability of individuals to understand and predict others' mental states, including their intentions and goals for optimal decision-making and efficient social collaboration. Recent work introduces the ToM module in single-agent scenarios to predict an agent's future actions based on its historical trajectory \cite{rabinowitz2018machine}. A ToM-based multi-agent communication method incorporates an additional ToM module to select optimal communication partners based on historical trajectories in multi-agent systems \cite{wang2021tom2c}. The mutual ToM process, introduced in recent work \cite{zhang2024mutual}, enhances human-AI collaboration by improving reasoning and communication. 
\paragraph{Partner modeling.} In multi-agent reinforcement learning (MARL) tasks, partners often exhibit dynamic and diverse strategies, introducing significant challenges in non-stationarity and generalization. To address these challenges in multi-agent collaboration, several studies~\cite{carroll2019utility,shih2021critical} have explored partner modeling to predict partners' behaviors for enhancing efficient collaboration in complex MARL scenarios. Other approaches construct additional behavioral models~\cite{li17mixture,jiang2024learning} to simulate human partners and characterize their behavior styles. However, these methods focus on the accuracy of behavioral policy modeling, neglecting the simulation of partners' mental characteristics. Our work focuses on human partner's multi-scale ToM modeling for partner knowledge, cognitive styles, and domain intentions. 

\begin{figure*}[htb]
\begin{center}
\includegraphics[width=0.96\textwidth]{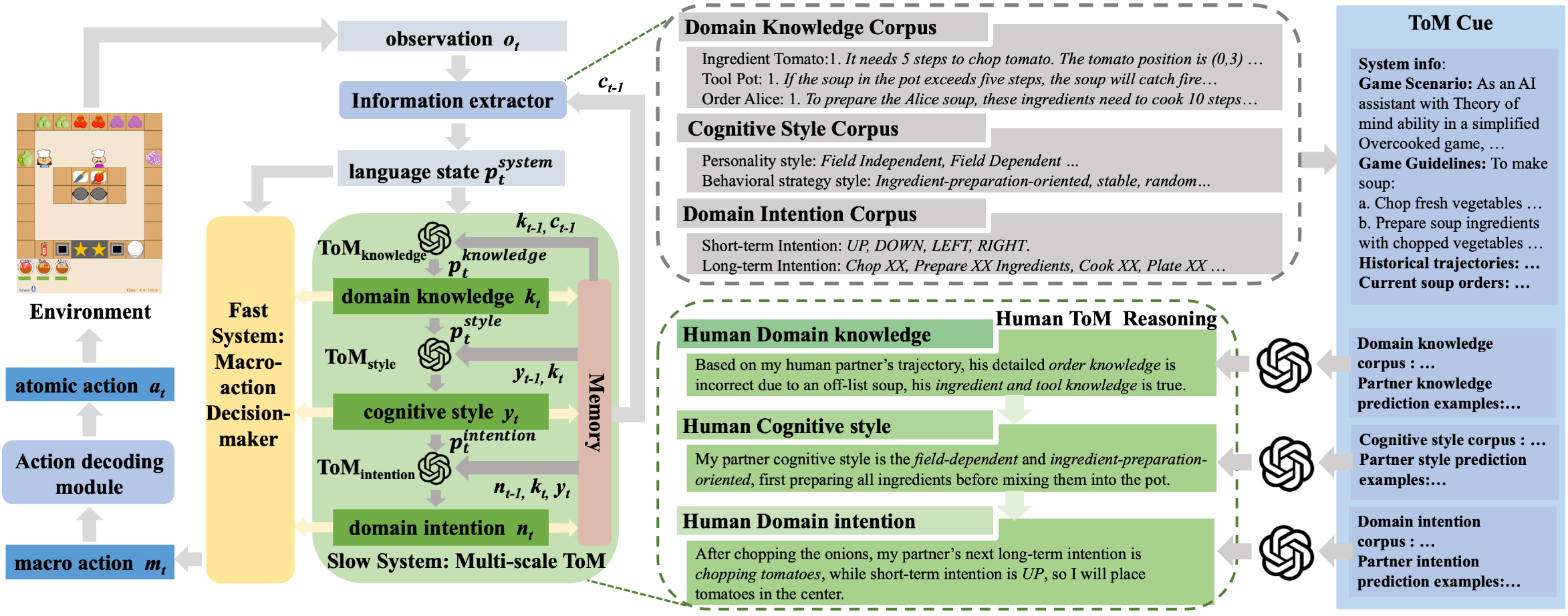}
\end{center}
\caption{As shown in the figure above, our proposed DPMT framework is inspired by the dual process theory, comprising an information extractor module, an action decoding module, a fast system for macro-action decision-making and a slow system for human partner modeling with a multi-scale ToM module. The ToM module consists of three stages: domain knowledge reasoning $\textrm{ToM}_\textrm{knowledge}$, cognitive style reasoning $\textrm{ToM}_\textrm{style}$, and domain intention reasoning $\textrm{ToM}_\textrm{intention}$. By emulating the cognitive structure, our framework can effectively reason about the human partner and achieve efficient collaboration.}
\label{mainsystem}
\end{figure*}
\section{DPMT: Dual Process Multi-scale Theory of Mind Framework}
\subsection{Overview}

As illustrated in Figure~\ref{mainsystem}, our DPMT method comprises several components: the information extractor, the slow reasoning system, the fast intuitive system, and the action decoding module. The information extractor transforms environmental observation $o_t$ and the human partner’s trajectory $c_{t-1}$ into the language state, represented as the system cue $p_t^\textrm{system}$. $p_t^\textrm{system}$ encodes textual information about the environment and serves as the input for both the fast and slow systems. 

The fast system focuses on the quick intuitive decision-making for each step, making macro-action $m_{t}$ from a predefined macro-action set. In contrast, the slow system emphasizes cognitive multi-scale ToM reasoning, modeling the partner's mental characteristics $k_{t}, y_{t},n_{t}$ to assist the fast system in making macro-action decisions.

The action decoding module decomposes the current macro-action $m_{t}$ into atomic actions $a_{t}$ and executes them at a much higher frequency until $m_{t}$ is completed. Once $m_{t}$ is finished, the fast system determines the subsequent $m_{t+1}$ based on the multi-scale partner reasoning $k_{t},y_{t},n_{t}$ from the slow reasoning system. This hierarchical approach ensures a seamless integration of intuitive decision-making and cognitive reasoning for efficient human-AI collaboration.

\subsection{Slow System}
\begin{figure}[!h]
\begin{center}
\includegraphics[width=0.48\textwidth]{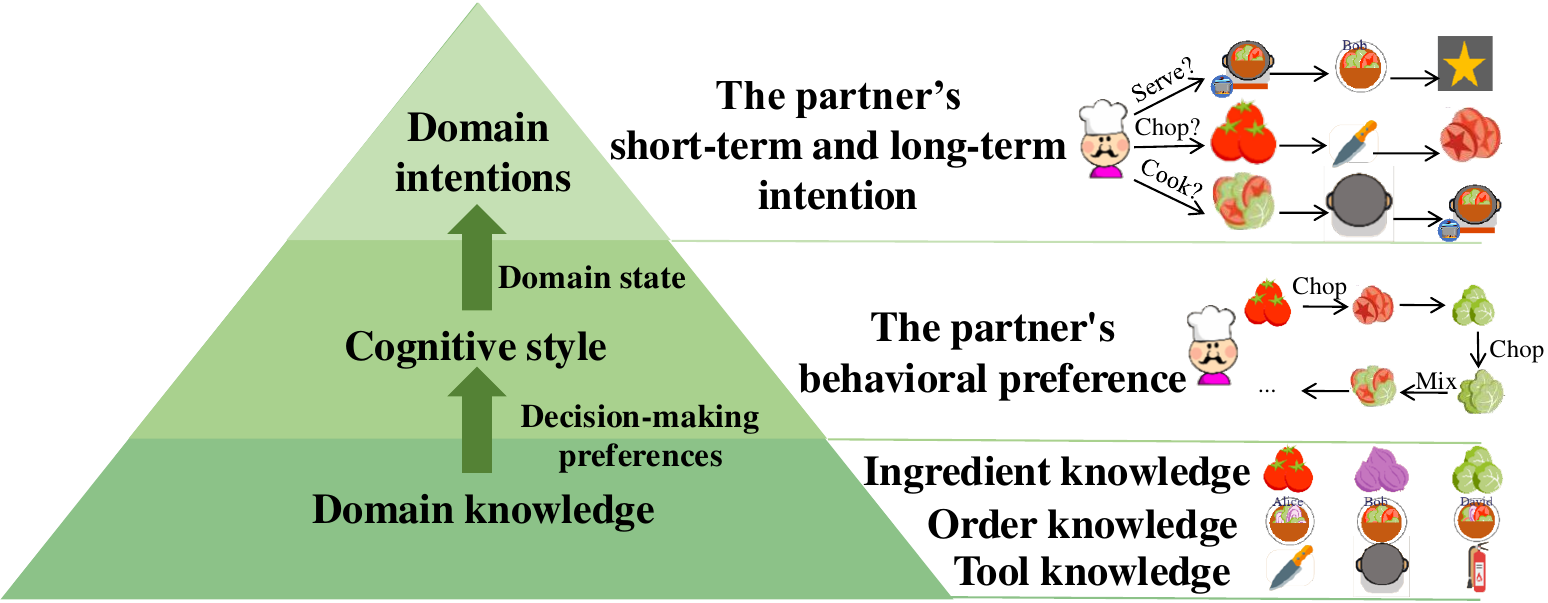}
\end{center}
\caption{Our proposed multi-scale mental characteristics system includes domain knowledge, cognitive style, and domain intention. Domain knowledge includes ingredient knowledge, order knowledge, and tool knowledge. Cognitive style represents the partner's personality trait and behavioral style. Domain intention consists of long-term and short-term goals, representing the partner’s decision-making process.}
\label{mentalsystem}
\end{figure}

As illustrated in Figure~\ref{mentalsystem}, we propose a multi-scale mental characteristics system for human-AI collaboration, drawing inspiration from social psychology research on human decision-making \cite{hauser2019social, santos2008social}. These studies categorize various mental characteristics that influence individual behavior into three key dimensions: \textit{domain knowledge}, \textit{cognitive style}, and \textit{domain intention} \cite{zhao2015role, bostan2009player}. This system organizes these mental characteristics into a hierarchical architecture, with domain knowledge forming the foundation and domain intentions at the top, progressively enhancing interpretability. Building on this, we propose a multi-scale ToM model as our slow system, simulating the slow cognitive process in the dual process theory. This ToM model comprises multiple stages of human partner mental characteristic reasoning: the human domain knowledge reasoning stage $\textrm{ToM}_\textrm{knowledge}$, the human cognitive style reasoning stage $\textrm{ToM}_\textrm{style}$, and the human domain intention reasoning stage $\textrm{ToM}_\textrm{intention}$.

Small models face significant limitations in interpretability and general knowledge, making it challenging to construct an efficient and interpretable ToM model capable of effectively capturing partner styles and inferring domain intentions from partner trajectories. To overcome these challenges, we leverage LLMs as the foundation for the ToM model, leveraging their strengths in interpretability and extensive world knowledge to enhance human partner ToM modeling. 
\subsubsection{Domain knowledge reasoning stage}  Domain knowledge in the \textit{Overcooked} includes \textit{ingredient knowledge}, \textit{order knowledge}, and \textit{tool knowledge}. Since these types of knowledge are not directly provided in the environmental state, they are typically accumulated from the human partner's experiences. Predicting human partners' domain knowledge serves as the foundation for further predictions of their cognitive styles and domain intentions, forming a critical basis for higher-level ToM reasoning. 

In our framework, the partner domain knowledge reasoning stage $\textrm{ToM}_\textrm{knowledge}$ takes the knowledge cue $p_t^\textrm{knowledge}$ as input, which integrates information from the language state $p_t^\textrm{system}$, the human partner’s trajectory context $c_{t-1}$, additional \textit{partner domain knowledge reasoning cases}, and a \textit{customized domain knowledge corpus}. The \textit{domain knowledge corpus} comprises knowledge from three categories, such as \textit{ingredient tomato} under \textit{ingredient knowledge}, \textit{order Bob} under \textit{order knowledge}. The output of $\textrm{ToM}_\textrm{knowledge}$ is the human partner domain knowledge $k_t$, which includes insights such as whether the human partner understands the current task rules or the required order preparation sequences.

\subsubsection{Cognitive style reasoning stage}
The cognitive style reflects a human partner’s mental characteristic for decision-making preferences based on the domain knowledge, including \textit{personality traits}, \textit{behavioral strategy preferences}, and \textit{risk preferences}. From the perspective of \textit{personality traits}, cognitive style can be categorized into \textit{field-independent}, who prefer to complete an entire order independently, and \textit{field-dependent}, who tend to collaborate by dividing order tasks into sub-tasks to complete an order. In field-dependent players, cognitive styles can be further categorized based on the behavioral strategy, which defines different task tendencies. For example, an \textit{ingredient-preparation-oriented} style focuses primarily on retrieving and processing various ingredients. This style can also be divided into stable (consistently following a fixed strategy) and random according to the behavioral strategy. Accurate classification of cognitive styles is crucial for effective partner intention modeling.

In the partner style reasoning stage $\textrm{ToM}_\textrm{style}$, the cognitive style cue $p_t^\textrm{style}$ comprises $p_t^\textrm{knowledge}$, the previous cognitive style prediction $y_{t-1}$, and the partner domain knowledge prediction $k_t$. This stage also incorporates an \textit{additional partner style corpus} to predict the human partner’s cognitive style $y_t$. In the partner style corpus, we design various partner cognitive styles based on behavioral strategies and personality traits observed in human behavioral experiments. These partner styles are stored in a structured format consisting of a name, a descriptive paragraph, and representative cases.

\subsubsection{Domain intentions reasoning stage}
Building upon the cognitive style and the current domain state, domain intentions include both \textit{short-term and long-term intentions} of the human partner, which are crucial for achieving efficient human-AI collaboration. \textit{Short-term intention} reasoning involves atomic action prediction, determining the human partner’s current action, such as \textit{UP}, \textit{DOWN}, \textit{LEFT}, \textit{RIGHT}. In contrast, \textit{long-term intention} reasoning focuses on predicting the human partner’s macro-action, specifying their primary goal for the current phase, such as \textit{Chop Tomato}, \textit{Prepare Bob Ingredients}, \textit{Cook Alice Soup}, \textit{Plate David Soup}. Based on the $k_t$ and $y_t$, the domain intention reasoning stage $\textrm{ToM}_\textrm{intention}$ incorporates a \textit{short-term intention cases} and \textit{long-term intention reasoning corpus} in domain intention cue $p_t^\textrm{intention}$ to output the human partner’s predicted intentions $n_{t}$.

\subsection{Fast System}
To enable more real-time human-AI collaboration, our DPMT incorporates a fast system that simulates the quick decision-making system in the dual process theory. While powerful large models like GPT-4o demonstrate superior reasoning ability, their high latency makes them unsuitable for the fast system. Instead, our fast system in DPMT leverages a smaller-scale LLM, such as llama-13B, to reduce latency in macro-action decision-making and achieve real-time collaboration, using the environmental language state $p_t^{system}$, the fast system prompt, and currently available actions as input. 

\begin{figure}[htb]
\begin{center}
\includegraphics[width=0.48\textwidth]{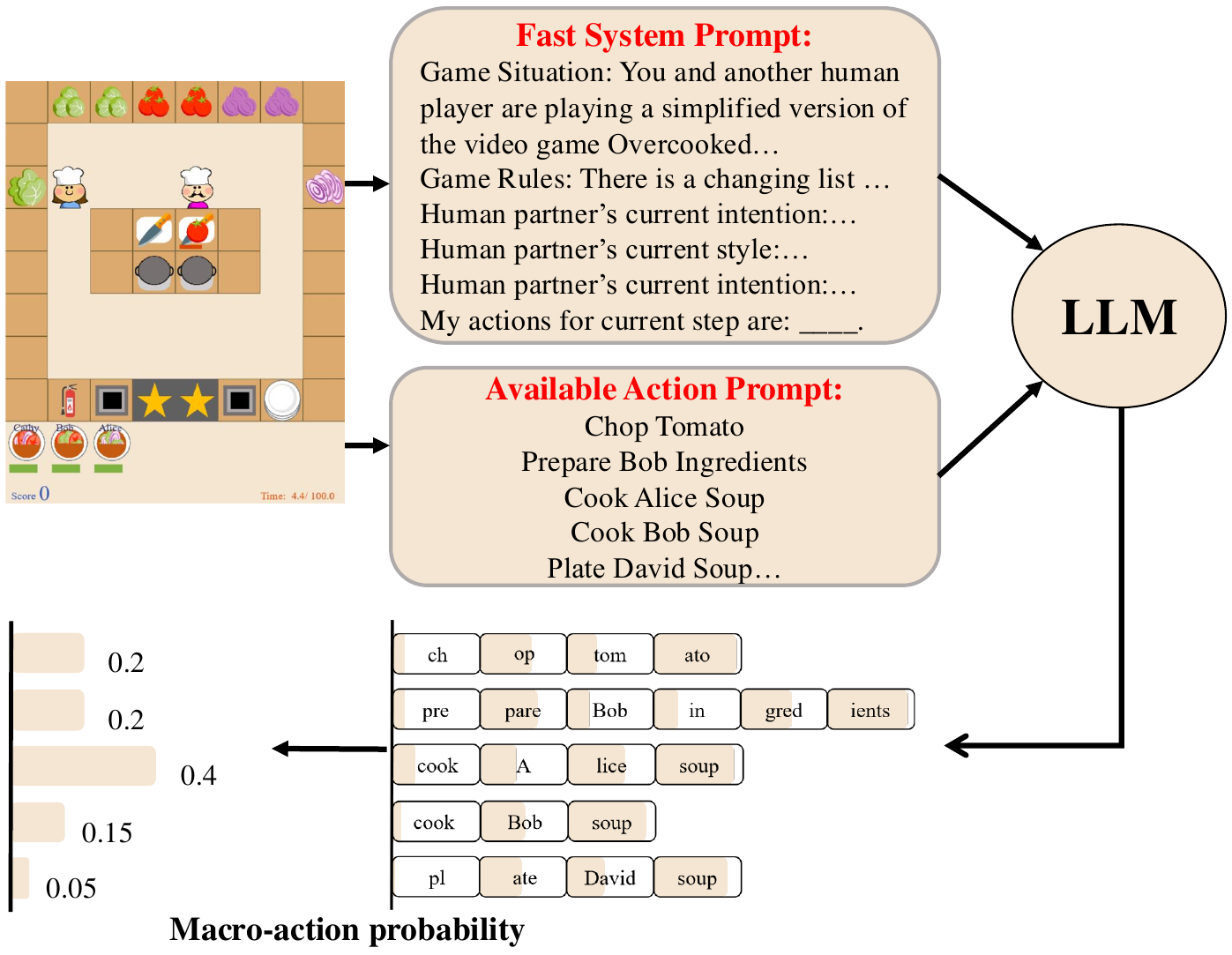}
\end{center}
\caption{Overview of our fast system in our proposed DPMT. The fast system first computes token probabilities for every macro-action, then derives the probability distribution of available macro-actions, selecting the one with the highest probability as the output.}
\label{quicksystem}
\end{figure}

As illustrated in Figure~\ref{quicksystem}, the fast system calculates macro-action probabilities by computing token probabilities, similar to recent works~\cite{liu2023llm,tantrue}. Each macro-action is represented as a sequence of tokens. The LLM evaluates the probability of token sequences based on the human ToM reasoning from the slow system. These probabilities are then normalized to form the macro-action probability distribution at step $t$. The fast system selects the macro-action $m_t$ with the highest probability from this distribution.

Although smaller models may make suboptimal decisions, our fast system compensates for this by leveraging accurate partner reasoning from the slow system. These two systems are implemented using multi-threading, where the slow system’s reasoning for partner modeling operates at a longer time scale than the fast system’s macro-action decisions. This design allows the fast system to run more frequently, enabling efficient real-time human-AI collaboration.
 
\section{Experiments}
\subsection{Experimental Environment}
\begin{figure}[htb]
\begin{center}
\includegraphics[width=0.48\textwidth]{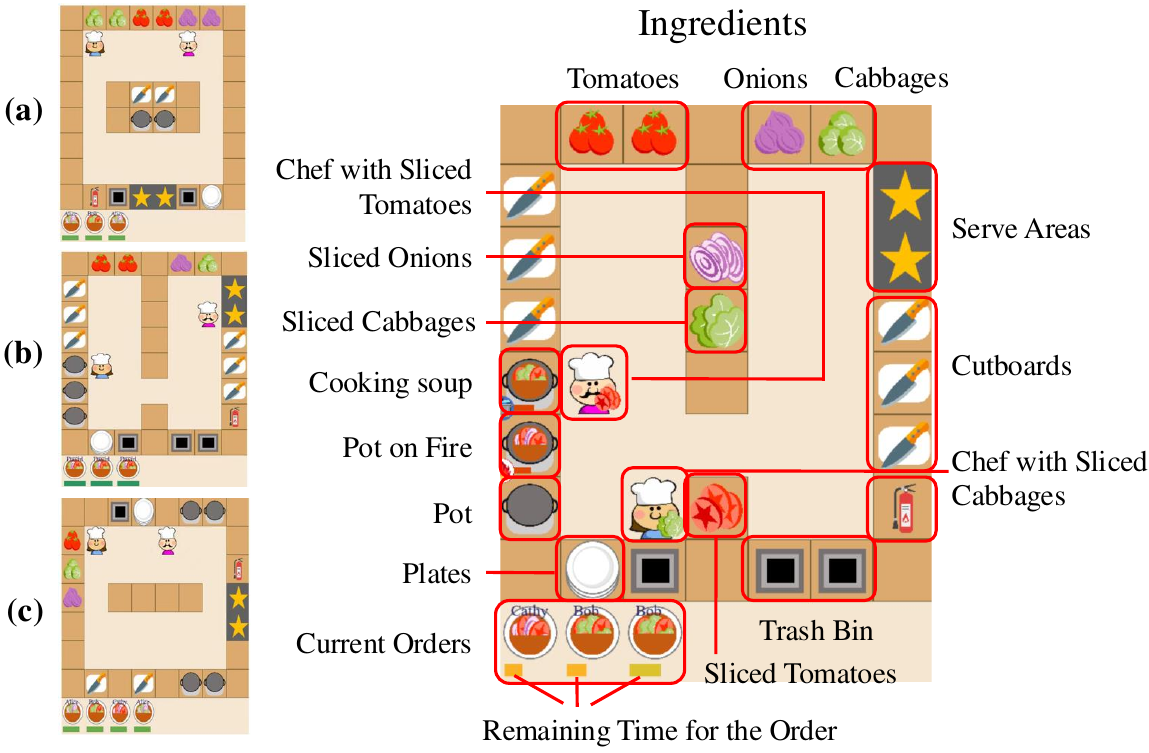}
\end{center}
\caption{The left part shows our Overcooked environments. The maps from top to bottom are Ring, Bottleneck, and Quick. The right part shows the detailed layout description of the bottleneck environment.}
\label{env-figure}
\end{figure}
Our experimental environment is \textit{Overcooked} \cite{ghosttown2016overcooked,liu2023llm}, a primary human-AI collaboration benchmark. We have conducted experiments on three maps based on the recent work \cite{liu2023llm}, which introduced an expanded version of the traditional \textit{Overcooked} environment for human-AI collaboration. In this environment, the agent and the human player work together to complete cooking tasks as quickly as possible to achieve a higher reward within a limited time. As shown in Figure~\ref{env-figure}, the environment includes three maps: \textit{Ring}, \textit{Bottleneck}, and \textit{Quick}, each with varying levels of difficulty. Among them, the \textit{Quick} and \textit{Bottleneck} maps are more challenging. As shown in Figure~\ref{env-figure}, completing an order requires following a specific sequence of steps. The agent and human player must retrieve the necessary ingredients, chop them, combine them according to the different recipes, cook them for the required duration to make a dish, plate the dish before it overcooks, and serve the dish to the designated position. Once the dish is served, the player receives a reward (for example, 15 points), while failing to submit a dish results in a penalty (for example, -5 points). Players can choose and execute one action simultaneously from the following options: UP, DOWN, LEFT, or RIGHT, using the directional arrow keys. This environment designs a macro-action set with 21 predefined macro actions for the LLM agent, such as \textit{Cook Alice Soup}.

\subsection{Human Experiment 1: Collaborating with Human Partners with Specific Strategies}
\subsubsection{Experimental Setup.}In the human-AI collaboration experiment, similar to HLAgent \cite{liu2023llm}, the human player followed the fixed strategy to control the partner agent and interact with objects using the arrow keys like 'up,' and 'down.' In our human experiments, baseline methods included our DPMT with Qwen, our DPMT without the multi-scale ToM (MsToM) and the HLAgent. Since communication between agents was not allowed in our task setting, we used the 0-command HLAgent as our baseline. To ensure experimental fairness, we established a stopping criterion that terminated LLM agent executions once the trajectory length reached 500 steps. In this part of the experiment, we used game scores to measure human-AI collaboration efficiency. The experiments were conducted on three maps and repeated five times under consistent conditions. The average human-AI collaboration scores are shown in Table~\ref{tab1}. 

\begin{table}[htbp]
\begin{center}
\caption{Score comparison between baseline methods and our DPMT method across all three maps in the human experiment 1. Standard deviations are shown in parentheses.}
\label{tab1}
\vskip 0.12in
\fontsize{7}{10}\selectfont
        \begin{tabular}{c|ccc}
        \hline
        Methods & \textit{Ring} & \textit{Bottleneck} & \textit{Quick}     \\
        \hline
        Our DPMT  & \textbf{121~($\pm$13.56)}  & \textbf{101~($\pm$13.56)} & 104~($\pm$16.73)    \\
        Our DPMT w Qwen & 100~($\pm$15.17) & 95~($\pm$22.80) & \textbf{105~($\pm$18.71)}   \\
        Our DPMT w/o MsToM  & 44~($\pm$22.23) & 23~($\pm$22.93) & 9~($\pm$19.34)   \\
        HLAgent  & 99~($\pm$18.81) & 60~($\pm$20.74) & 87~($\pm$16.00)   \\
        
        \hline
        \end{tabular}
\end{center}
\end{table}
\subsubsection{Main Results.} As shown in Table~\ref{tab1}, our proposed DPMT method consistently achieved the highest collaboration performance when paired with fixed-strategy human partners across all three maps: \textit{Ring}, \textit{Quick}, and \textit{Bottleneck}, highlighting the effectiveness of our DPMT framework. Compared to the baseline methods, the introduction of the multi-scale mental characteristic ToM reasoning module (MsToM) in our DPMT framework led to a higher success rate in serving soups, a lower obstruction rate, and a lower incidence of dish fires by accurately predicting human partners' mental characteristics, such as domain intentions. In contrast, the lack of ToM reasoning in DPMT w/o MsToM often resulted in ineffective collaboration, including redundant actions, route blockages, overcooked dishes, and frequent task repetition, ultimately incurring performance penalties.

\begin{figure}[htb]
\begin{center}
\includegraphics[width=0.48\textwidth]{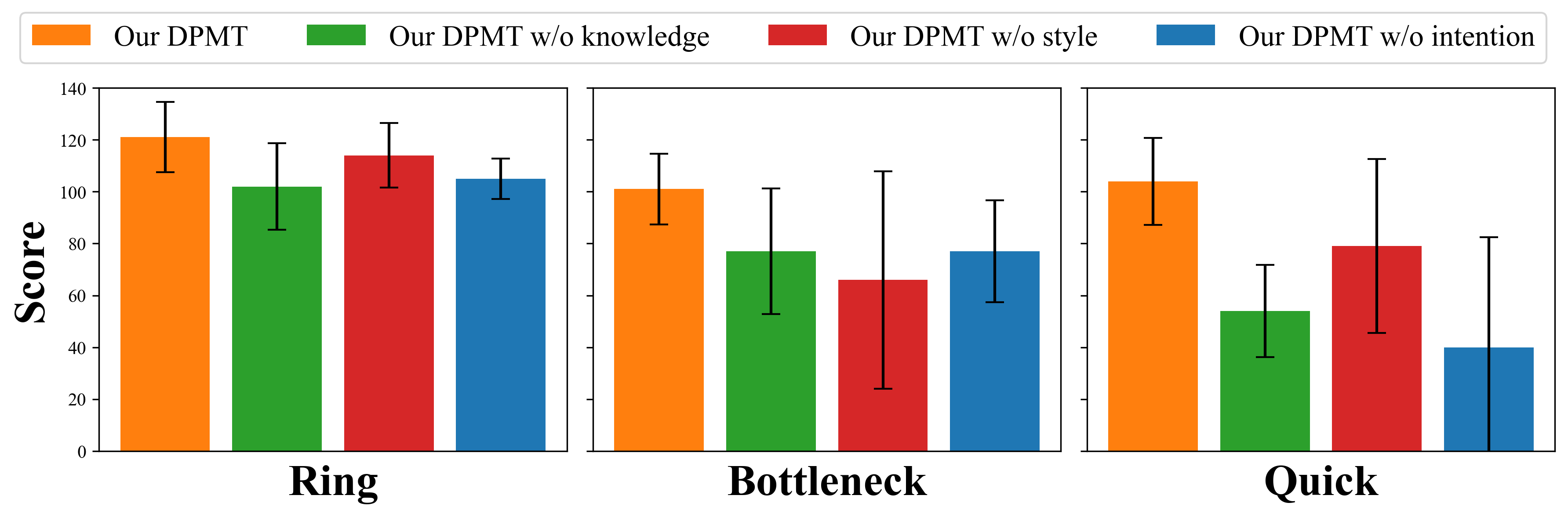}
\end{center}
\caption{Score results of ablation experiments for different mental characteristics in human experiment 1. The black line denotes the standard deviation.}
\label{result3}
\end{figure}
Additionally, compared to the HLAgent baseline, our method achieved superior and more stable performance, particularly on complex maps such as \textit{Bottleneck}. The \textit{Bottleneck} map has a more complex structure with a narrow passage, resulting in a higher blockage rate. This highlights the critical role of accurate ToM reasoning in achieving efficient human-AI collaboration. Furthermore, substituting the GPT-4o API with Qwen-72B as the base model for mental characteristic reasoning resulted in slight improvements in collaboration performance on the \textit{Quick} map. This enhancement is likely attributed to minor fluctuations in real-time experimental results caused by latency.
\subsubsection{Ablation Results.}
We conducted ablation experiments to analyze the effectiveness of our DPMT method further. As shown in Figure~\ref{result3}, we compared the complete DPMT method with its ablated versions, where specific mental characteristic reasoning stages (w/o knowledge, w/o style, and w/o intention) were removed. Our method consistently outperformed the ablated versions across all maps (\textit{Ring}, \textit{Quick}, and \textit{Bottleneck}), especially on more complex maps, where effective ToM reasoning plays a crucial role in coordination. For example, removing the intention reasoning layer led to a significant performance drop on the \textit{Quick} map, while omitting the style layer caused notable declines on the \textit{Bottleneck} map. These results indicate the importance of each ToM characteristic and demonstrate the necessity of hierarchical ToM modeling for effective partner reasoning and improved collaboration.

\begin{figure}[htb]
\begin{center}
\includegraphics[width=0.45\textwidth]{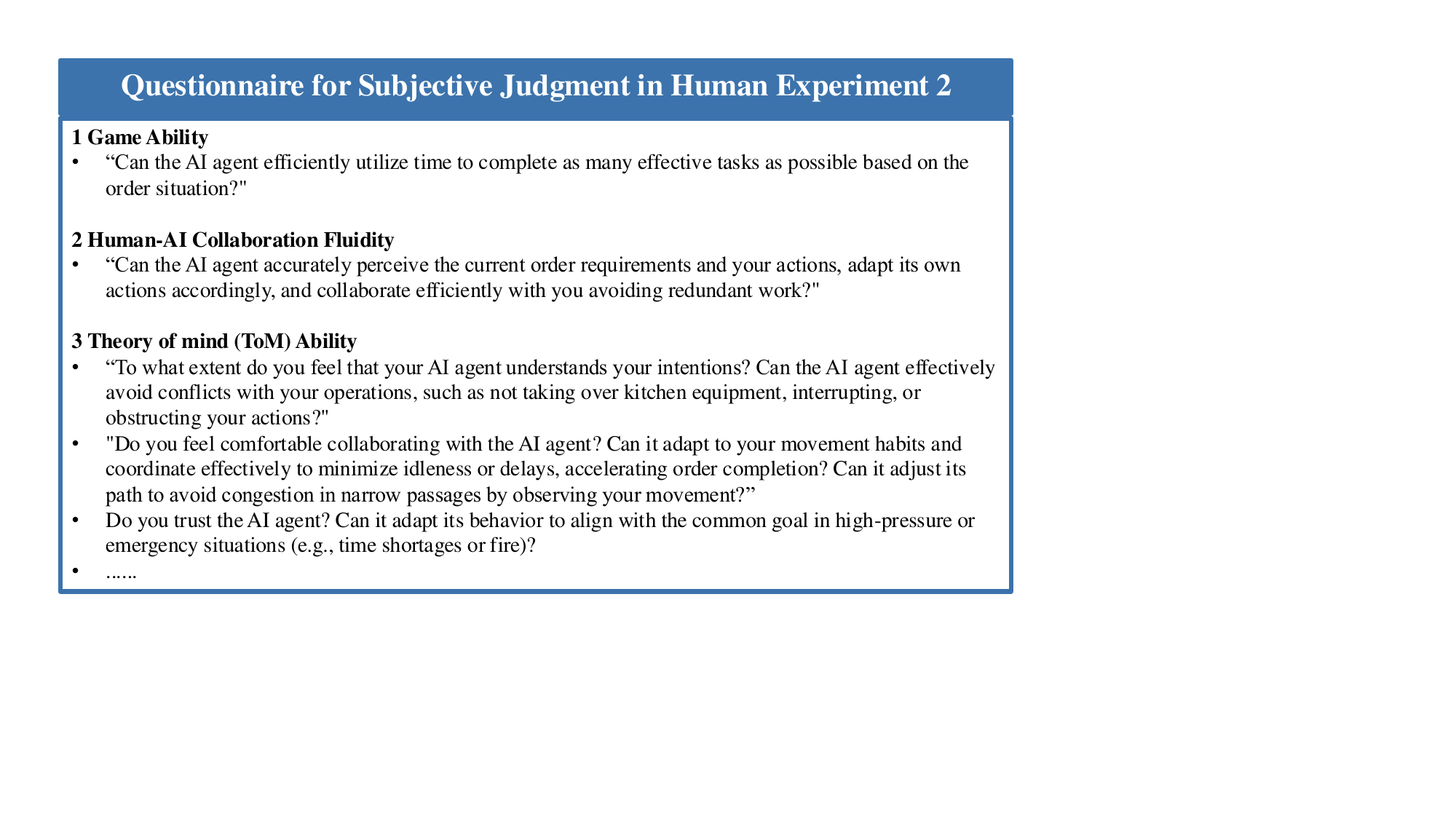}
\end{center}
\caption{Our proposed questionnaire for subjective judgment metric with 5-point Likert scales was used in our human experiment 2. The questionnaire includes three parts: game ability, human-AI collaboration fluidity, and ToM ability. Different questionnaires were designed for different maps.}
\label{mentalq}
\end{figure}
\subsection{Human Experiment 2: Collaborating with Human Partners with Diverse Strategies}
\subsubsection{Experiment Setup.} An ideal AI agent should not only cooperate seamlessly with fixed-strategy human players but also collaborate efficiently with unseen human partners who employ diverse strategies. To further analyze the effectiveness and the human preference of our proposed DPMT framework, we conducted additional behavioral experiments for human studies. These experiments were performed across three maps with 30 participants (26 male, 4 female, ages 20–50). As shown in Figure~\ref{mentalq}, we designed a 5-point Likert scale questionnaire for each map, consisting of several questions  \cite{hoffman2019evaluating,zhang2024mutual}.  This experiment followed a within-subject design, where each participant collaborated with both the DPMT and three baseline agents (A-D). To ensure fairness and validity, we randomized the order of partner agents ('ABCD' or 'DBCA') for each participant. Additionally, to evaluate the AI agent’s performance more effectively, we extended the game duration to 90 seconds, longer than in previous human experiment 1.

All participants were required to review the informed consent form and game rules before starting. To familiarize themselves with the task environment and rules, they first collaborated with rule-based agents across three maps. After completing the collaboration task on each map, they filled out a corresponding questionnaire, rating different agents based on their level of agreement with each statement.

\begin{figure}[htb]
\begin{center}
\includegraphics[width=0.48\textwidth]{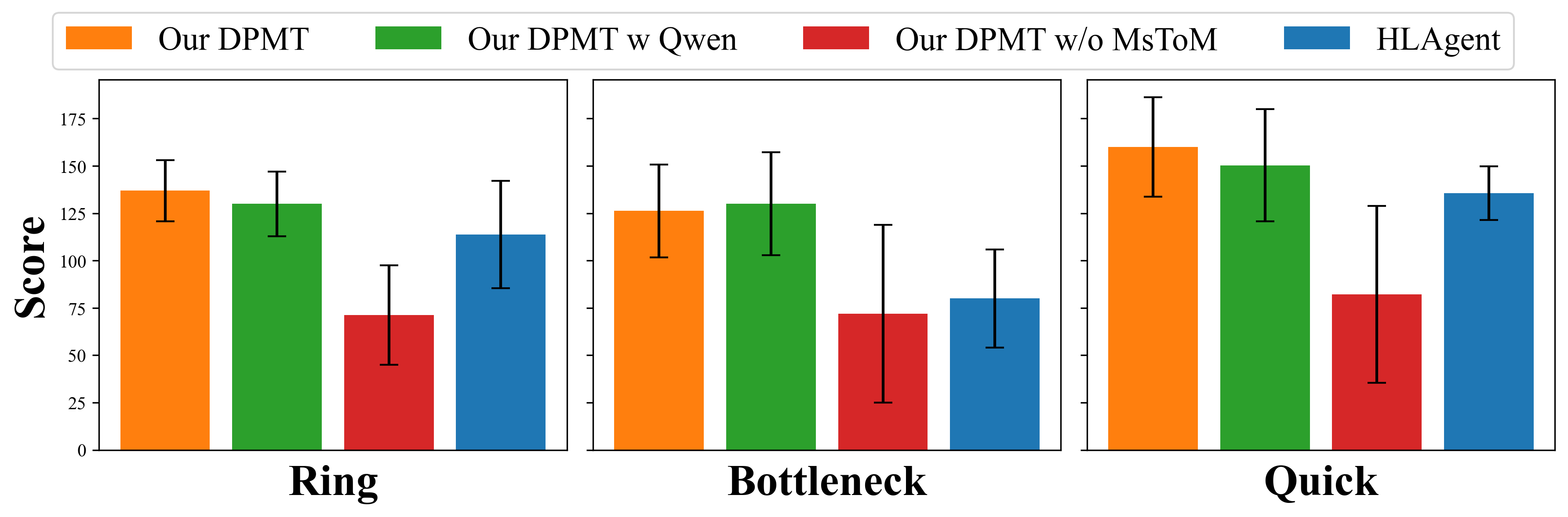}
\end{center}
\caption{Average collaboration score comparison between baseline methods and our proposed DPMT when paired with diverse participants in the human experiment 2.}
\label{result2}
\end{figure}
\subsubsection{Results.}
In our collaboration experiments with diverse human participants, we evaluated performance using two key metrics: average collaboration scores within a fixed time limit and subjective judgments from the participants. As shown in Figure~\ref{result2}, our DPMT method consistently outperformed baseline methods across all three maps, exhibiting a result trend similar to the main result of human experiment 1, demonstrating its ability to adapt to both specific strategies and diverse unseen human collaborators. For the subjective judgment scores, we computed the average scores based on our proposed questionnaire, with a maximum score of 5, as shown in Table~\ref{resulthuman}. Notably, our method achieved higher subjective scores, further highlighting its superior ToM reasoning capability for effective partner modeling.

\begin{table}[htbp]
\begin{center}
\caption{The average subjective judgment when paired with diverse unseen participants in human experiment 2 ranges from 0 to 5. Standard deviations are shown in parentheses.}
\label{resulthuman}
\vskip 0.12in
\fontsize{7}{10}\selectfont
        \begin{tabular}{c|ccc}
        \hline
        Methods & \textit{Ring} & \textit{Bottleneck} & \textit{Quick}     \\
        \hline
        Our DPMT  & \textbf{4.07~($\pm$0.53)}  & \textbf{4.02~($\pm$0.45)} & \textbf{4.02~($\pm$0.61)}    \\
        Our DPMT w Qwen & 3.51~($\pm$0.58) & 3.73~($\pm$0.47) & 3.89~($\pm$0.68)   \\
        Our DPMT w/o MsToM  & 2.20~($\pm$0.70) & 2.16~($\pm$0.67) & 2.04~($\pm$0.60)   \\
        HLAgent  & 3.09~($\pm$0.30) & 2.78~($\pm$0.59) & 3.22~($\pm$0.69)   \\
        \hline
        \end{tabular}%
\end{center}
\end{table}

\section{Conclusion}
Drawing inspiration from the dual process theory, we introduce a cognitive dual process multi-scale ToM (DPMT) framework, which simulates both the fast system for quick decision-making and the slow system for ToM reasoning. The slow system in DPMT employs a three-stage ToM reasoning process—partner domain knowledge, cognitive style, and domain intention—to improve the interpretability and generalization of partner ToM modeling. Additionally, it integrates a structured corpus of mental characteristics and a memory module that stores partner trajectories and ToM reasoning predictions. Our DPMT outperforms baseline methods in human-AI collaboration experiments, demonstrating superior adaptability and ToM ability. Ablation studies further validate the contribution of different reasoning stages. Notably, our method achieves significantly stronger results on complex maps, highlighting its potential for tackling more intricate human-AI collaborative tasks.

\section{Acknowledgments}
This work is supported by the Strategic Priority Research Program of the Chinese Academy of Sciences (No.XDA27040200) and the National Key R\&D Program of China
(No.2022ZD0116405).

\bibliographystyle{apacite}

\setlength{\bibleftmargin}{.125in}
\setlength{\bibindent}{-\bibleftmargin}

\bibliography{CogSci_Template}

\end{document}